\documentclass{article}
\usepackage{ijcai21}

\usepackage{times}
\usepackage{soul}
\usepackage{url}
\usepackage[hidelinks]{hyperref}
\usepackage[utf8]{inputenc}
\usepackage[small]{caption}
\usepackage{graphicx}
\usepackage{amsmath}
\usepackage{amsthm}
\usepackage{booktabs}
\usepackage{algorithm}
\usepackage{algorithmic}
\newtheorem{example}{Example}

\title{On Explaining Random Forests with SAT}

\author{
Yacine Izza$^1$
\and
Joao Marques-Silva$^2$
\affiliations
$^1$ANITI, Univ. Toulouse, France\\
$^2$IRIT, CNRS, Toulouse, France\\
\emails
yacine.izza@univ-toulouse.fr,
joao.marques-silva@irit.fr
}

\usepackage[table]{xcolor}
\usepackage{comment}
\usepackage{setspace}
\usepackage{empheq}
\usepackage{enumitem}
\usepackage{amssymb}
\usepackage{pgf}
\usepackage{tikz}
\usepackage{adjustbox}         
\usepackage{multirow,siunitx}
\usepackage{subfig}
\usepackage{titlesec}

\newtheorem{theorem}{Theorem}

\newtheorem*{remark*}{Remarks}

\newenvironment{Proof}{\noindent{\em Proof.~}}{\hfill$\Box$\\[-0.15cm]}

\newcommand{\fml}[1]{{\mathcal{#1}}}
\newcommand{\set}[1]{\{ #1 \}}

\newcommand{\tn}[1]{\textnormal{#1}}

\newcommand{\mbf}[1]{\ensuremath\mathbf{#1}}
\newcommand{\mbb}[1]{\ensuremath\mathbb{#1}}
\newcommand{\msf}[1]{\ensuremath\mathsf{#1}}

\newcommand{\ddp}{\tn{D}^{\tn{P}}}

\DeclareMathOperator*{\nentails}{\nvDash}
\DeclareMathOperator*{\entails}{\vDash}
\DeclareMathOperator*{\lequiv}{\leftrightarrow}
\DeclareMathOperator*{\limply}{\rightarrow}

\AtBeginDocument{

}

\begin{document}

\maketitle

\begin{abstract}
  Random Forests (RFs) are among the most widely used Machine Learning
  (ML) classifiers.
  Even though RFs are not interpretable, there are no dedicated
  non-heuristic approaches for computing explanations of RFs.
  Moreover, there is recent work on polynomial algorithms for
  explaining ML models, including naive Bayes classifiers.
  Hence, one question is whether finding explanations of RFs
  can be solved in polynomial time.
  This paper answers this question negatively, by proving that
  deciding whether a set of literals is a PI-explanation of an RF is
  $\ddp$-complete.
  Furthermore, the paper proposes a propositional encoding for
  computing explanations of RFs, thus enabling finding PI-explanations 
  with a  SAT
  solver. This contrasts with earlier work on explaining boosted trees
  (BTs) and neural networks (NNs), which requires encodings based on
  SMT/MILP.
  Experimental results, obtained on a wide range of publicly available
  datasets, demonstrate that the proposed SAT-based approach scales to
  RFs of sizes common in practical applications. Perhaps more
  importantly, the experimental results demonstrate that, for the vast
  majority of examples considered, the SAT-based approach proposed in
  this paper significantly outperforms existing heuristic approaches.
\end{abstract}

\section{Introduction}

The recent successes of Machine Learning (ML), and the forecast
continued growth of ML-enabled applications, including applications
that impact human beings or that are even safety critical, has raised
the need for identifying explanations for the predictions made by ML
models. As a result, recent years witnessed the rapid growth of the
field of explainable Artificial Intelligence (XAI) (see
e.g.\ \cite{GuidottiMRTGP19,LiLCR18,MontavonSM18,ShihCD18,ShihCD19,Ribeiro0G16,Ribeiro0G18,IgnatievNM19,Ignatiev20,marquis-kr20,inams-aiia20,DL:sat21a,DL:icml21}).
%
%
Unfortunately, the most promising ML models, including neural networks
or ensembles of classifiers, due to their size and intrinsic
complexity, are generally accepted to be non-interpretable (or
black-box), with the understanding that the predictions made by
such black-box models cannot be understood by human decision makers.

A large body of work on XAI is based on heuristic
approaches~\cite{Ribeiro0G16,LundbergL17,Ribeiro0G18}, offering
no formal guarantees regarding computed explanations\footnote{%
  For example, an explanation $E$, for an input $I_1$ resulting in
  prediction $A$, can also be consistent with input $I_2$ resulting in
  prediction $B\not=A$~\cite{Ignatiev20}.
  Such \emph{loose} explanations inevitably raise concerns in
  applications where safety is critical.}.
In contrast, recent work focused on non-heuristic approaches
which offer formal guarantees with respect to computed
explanations~\cite{ShihCD18,IgnatievNM19,ShihCD19,Ignatiev20,darwiche-ecai20,marquis-kr20,msgcin-nips20,inams-aiia20,DL:sat21a,DL:icml21}.

Approaches to explainability can also be characterized as being
model-agnostic or model-precise\footnote{%
  Orthogonal to the goals of the paper is the classification of
  explanations as \emph{local} or
  \emph{global}~\cite{GuidottiMRTGP19}.
  }.
Model-agnostic approaches do not require information about the ML
model representation, thus allowing the explanation of any class of ML 
models.
In contrast, in model-precise approaches, some representation of the
concrete ML model is reasoned about, and so these are characterized by
being model-specific. Whereas model-agnostic approaches are in general
heuristic, model-precise approaches can either be
non-heuristic~\cite{ShihCD18,IgnatievNM19,ShihCD19,Ignatiev20,darwiche-ecai20,marquis-kr20,msgcin-nips20,inams-aiia20,DL:sat21a,DL:icml21}
or heuristic~\cite{ZhaoWLC19,PetkovicAWV18,MollasBVT20}.
For model-precise non-heuristic approaches different solutions have 
been investigated.
\cite{ShihCD18}  propose an approach for explaining Bayesian network
classifiers, which is based on compiling such classifiers into Ordered
Decision Diagrams representing all prime implicants of the boolean
function representing the target class predictions.
These represent the so-called PI-explanations (which we revisit in
\autoref{sec:prelim}).
A different approach, based on abductive
reasoning~\cite{IgnatievNM19,Ignatiev20}, exploits automated
reasoning tools (e.g.\ SMT, MILP, etc.) with explanations being
computed on demand.
%
In abductive reasoning approaches, the ML model is represented as a
set of constraints and, given some target instance, a prime implicant
is computed, which represents a minimal set of feature-value pairs
that is sufficient for the prediction. Earlier work investigated
encodings of neural networks~\cite{IgnatievNM19} and of boosted
trees~\cite{Ignatiev20}.

This paper extends earlier work on model-precise non-heuristic
explainability. Concretely, the paper proposes a novel approach 
for computing PI (or abductive) explanations (AXps) of Random Forest classifiers
\cite{Breiman01,YangWJZ20,Zhang19,GaoZ20,FengZ18a,ZhouF17}.
Random Forests (RFs) represent a widely used tree ensemble ML model,
where each RF ML model is composed of a number of decision trees
(DTs).
(The importance of RFs is further illustrated by recent proposals
for implementing deep learning (DL) with
RFs~\cite{ZhouF17,Zhang19,FengZ18a}.)

In contrast with earlier work~\cite{Ignatiev20}, we show that in the
case of RFs it is possible to devise a purely propositional encoding.
In turn, this enables achieving very significant performance gains.
Concretely, the experimental results show that our approach is able to
compute explanations of realistically-sized RFs most often in a
fraction of a second. The experiments also show that our approach is
on average more than 
one order of magnitude faster
than a state of the art model-agnostic heuristic
approach~\cite{Ribeiro0G18}.

Recent work on model-precise non-heuristic explainability has shown
that some ML models can be explained in polynomial
time~\cite{marquis-kr20,msgcin-nips20}.
In contrast, this paper proves that it is $\ddp$-complete to decide
whether a set of literals is a PI-explanation (AXp) of an RF, thus making it
unlikely that RFs can be explained in polynomial time.

The paper is organized as follows.
%
\autoref{sec:prelim} covers the preliminaries.
\autoref{sec:cplx} proves the complexity of deciding whether a set of
literals is an explanation for an RF.
\autoref{sec:enc} proposes a propositional encoding for computing one
AXp of an RF.
\autoref{sec:res} presents the experimental results.
Finally, \autoref{sec:conc} concludes the paper.


\section{Preliminaries}
\label{sec:prelim}

\paragraph{ML Classification.}
We consider a machine learning classification problem, defined by a set of features 
$\fml{F}=\{1,\ldots,m\}$, and by a set of classes  $\fml{K} = \{c_1, c_2, \ldots, c_K \}$. 
Each feature $j \in \fml{F}$ takes values from a  domain $\mbb{D}_i$. 
(Domains may correspond to  Boolean, Categorical or Continuous data.) 
Thus, feature space is defined as 
$\mbb{F}=\mbb{D}_1\times{\mbb{D}_2}\times\ldots\times{\mbb{D}_m}$. 
To refer to an arbitrary point in feature space we use the notation
$\mbf{x}=(x_1,\ldots,x_m)$, whereas to refer to a specific point in
feature space we use the notation $\mbf{v}=(v_1,\ldots,v_m)$, with
$v_i\in{\mbb{D}_i}$, $i=1,\ldots,m$.
An \emph{instance} (or example) denotes a pair $(\mbf{v}, c)$, where 
$\mbf{v}\in\mbb{F}$ and $c\in\fml{K}$.
An ML classifier is characterized by a classification function $\tau$
that maps the feature space $\mbb{F}$ into the set of classes $\fml{K}$,
i.e.\ $\tau:\mbb{F}\to\fml{K}$.
To learn a classifier, a set of instances 
$\set{(\mbf{v}_1, c_1), \ldots, (\mbf{v}_k, c_k)}$ is used as training
data  by a learning algorithm that returns a function with a best fit
on the training data.

\paragraph{Decision Tree and Random Forest Classifiers.}
Decision trees rank among the most widely-used techniques ML
models~\cite{BreimanFOS84,Quinlan93}. 
Formally, a decision tree $\fml{T} = (V_{\fml{T}}, E_{\fml{T}})$  is a directed acyclic graph, 
where the root node has no incoming edges, and every other node has exactly 
one incoming edge. The nodes of a tree are partitioned into terminal ($V_{T}$) 
and non-terminal ($V_{NT}$) nodes. 
Terminal nodes denote the leaf nodes, and have no outgoing edges (i.e. children). 
Non-terminal nodes denote the internal nodes, and have outgoing edges. 
Each terminal node $j \in{V_T}$ is associated with a class taken from $\fml{K}$. 
We define a map $\kappa : V_T \to \fml{K}$ to represent the class associated 
with each terminal node. 
Each non-terminal node is associated with a feature taken from a set of features $\fml{F}$. 
Given a feature $j \in{\fml{F}}$ associated with a non-terminal node $l$, each outgoing edge 
represents a literal of the form  $x_j \bowtie S_j$, where either $S_j \in \mbb{D}_j$ or
$S_j \subseteq \mbb{D}_j$
\footnote{Features are either categorical (including boolean) or real-
  or integer-valued ordinal, and
  $\bowtie\:\in{\set{=,\in}}$. (Observe that these operators allow
  for intervals to be represented.)}.
Each path in $\fml{T}$ is associated with a consistent conjunction of literals, 
denoting the values assigned to the features so as to reach the terminal node in the path. 
We will represent the set of literals of some tree path $R_k$  by $\fml{L}(R_k)$.
A well-known drawback of decision trees is overfitting with respect to
the training data.
In contrast, tree ensembles such as Random Forests
(RFs)~\cite{Breiman01} combine several tree-based models, which allows
for improved accuracy and ability to generalize beyond the training
data.
More formally, an RF $\mathfrak{F}$ is a collection of decision trees
(DTs)  $\mathfrak{F} = \set{\fml{T}_1, \fml{T}_2, \ldots, \fml{T}_M }$. 
Each tree $\fml{T}_i \in{\mathfrak{F}}$ is trained on a subsample
of the training dataset  so as the trees of the RF are not correlated.
The prediction function in RF works by majority vote, that is 
each tree votes for a class and the most voted class is picked. (In
case of ties, for simplicitly we will pick the lexicographically
smallest class.)

\begin{figure}[t]
\centering
\begin{minipage}[b]{0.45\linewidth}
\centering
\subfloat[Tree 1] {
\includegraphics[height=3.5cm]{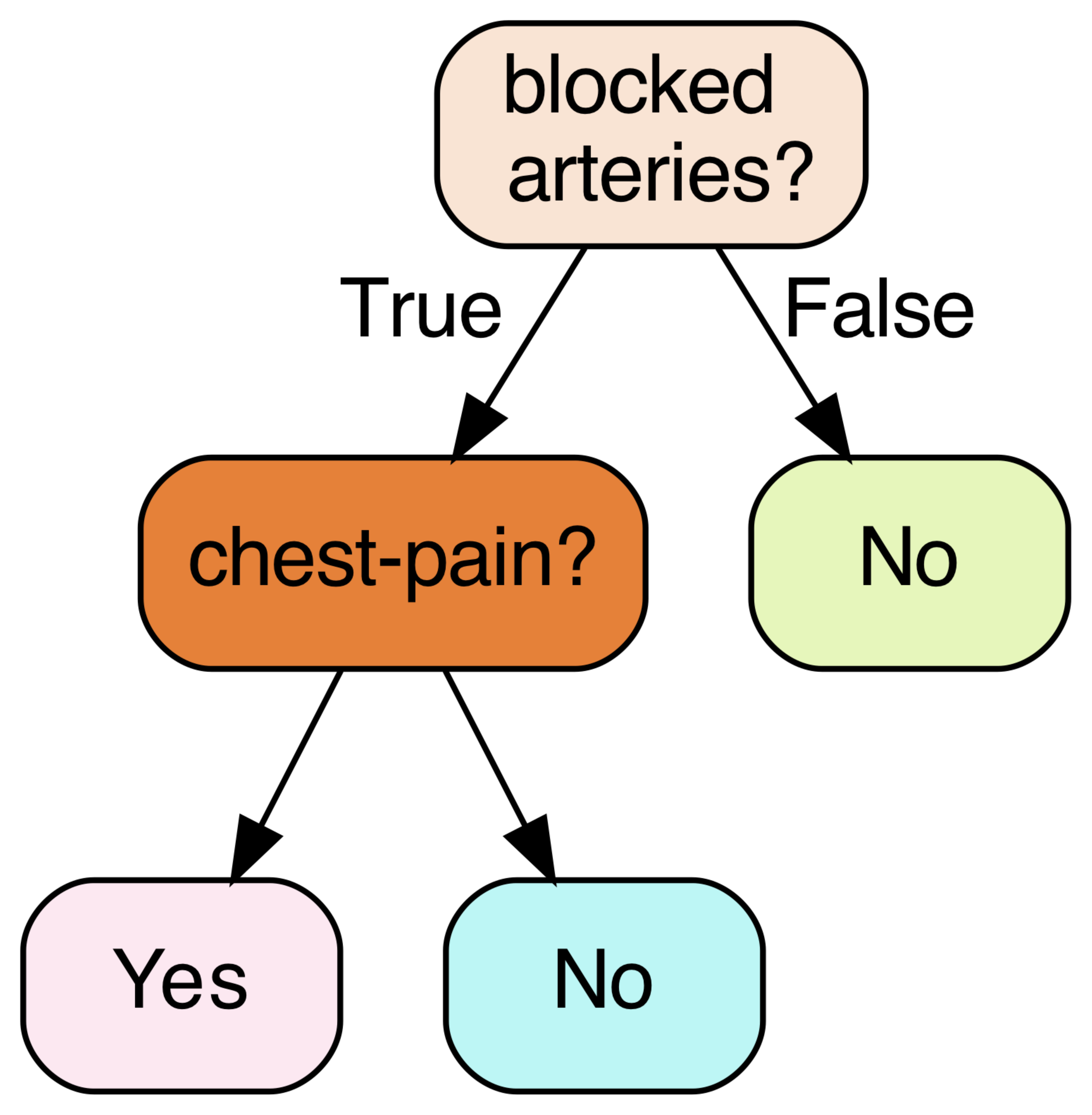}}
\end{minipage}
\begin{minipage}[b]{0.45\linewidth}
\centering
\subfloat[Tree 2] {
\includegraphics[height=3.5cm]{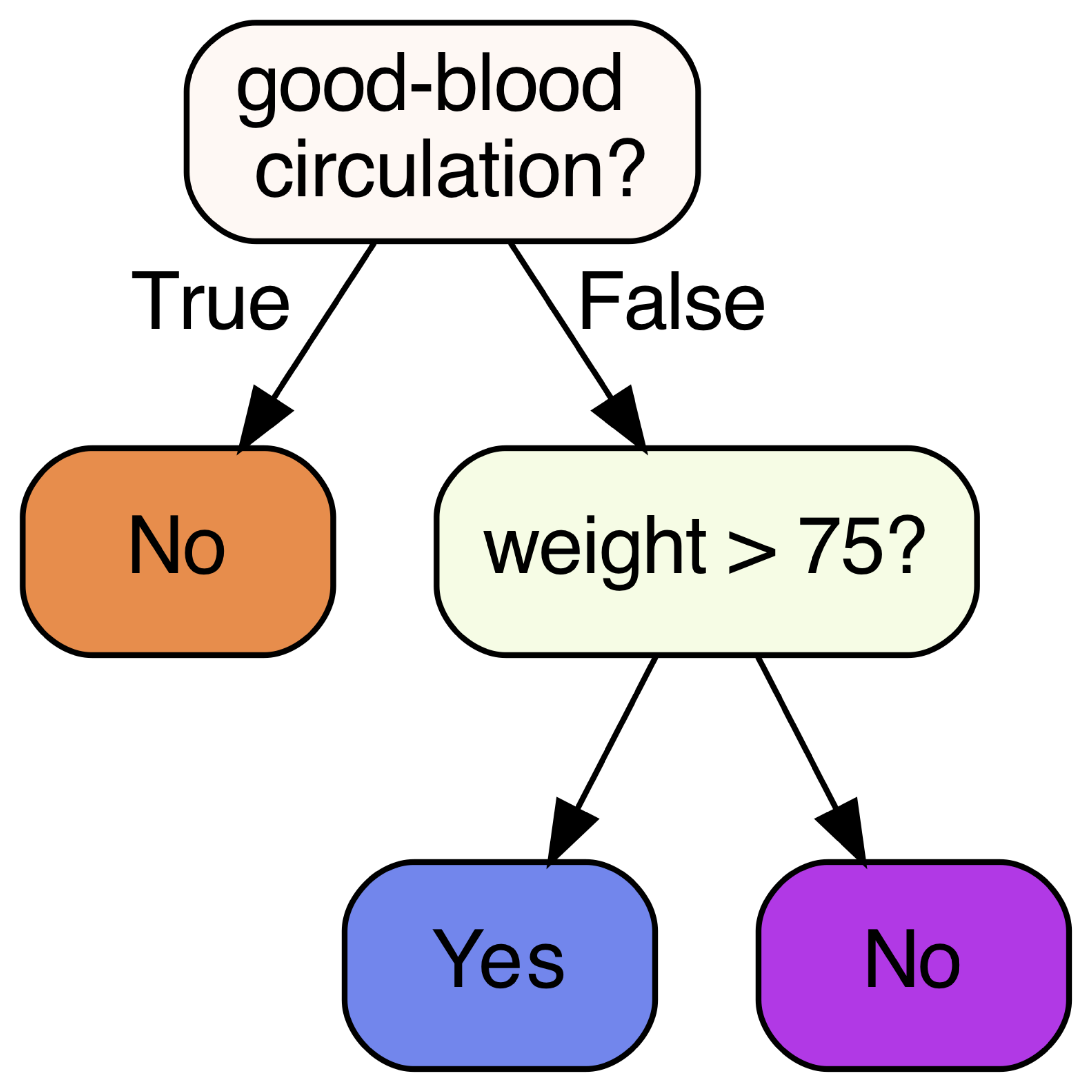}}
\end{minipage}
\\
\begin{minipage}[b]{\linewidth}
\centering
\subfloat[Tree 3] {
\includegraphics[height=3.5cm]{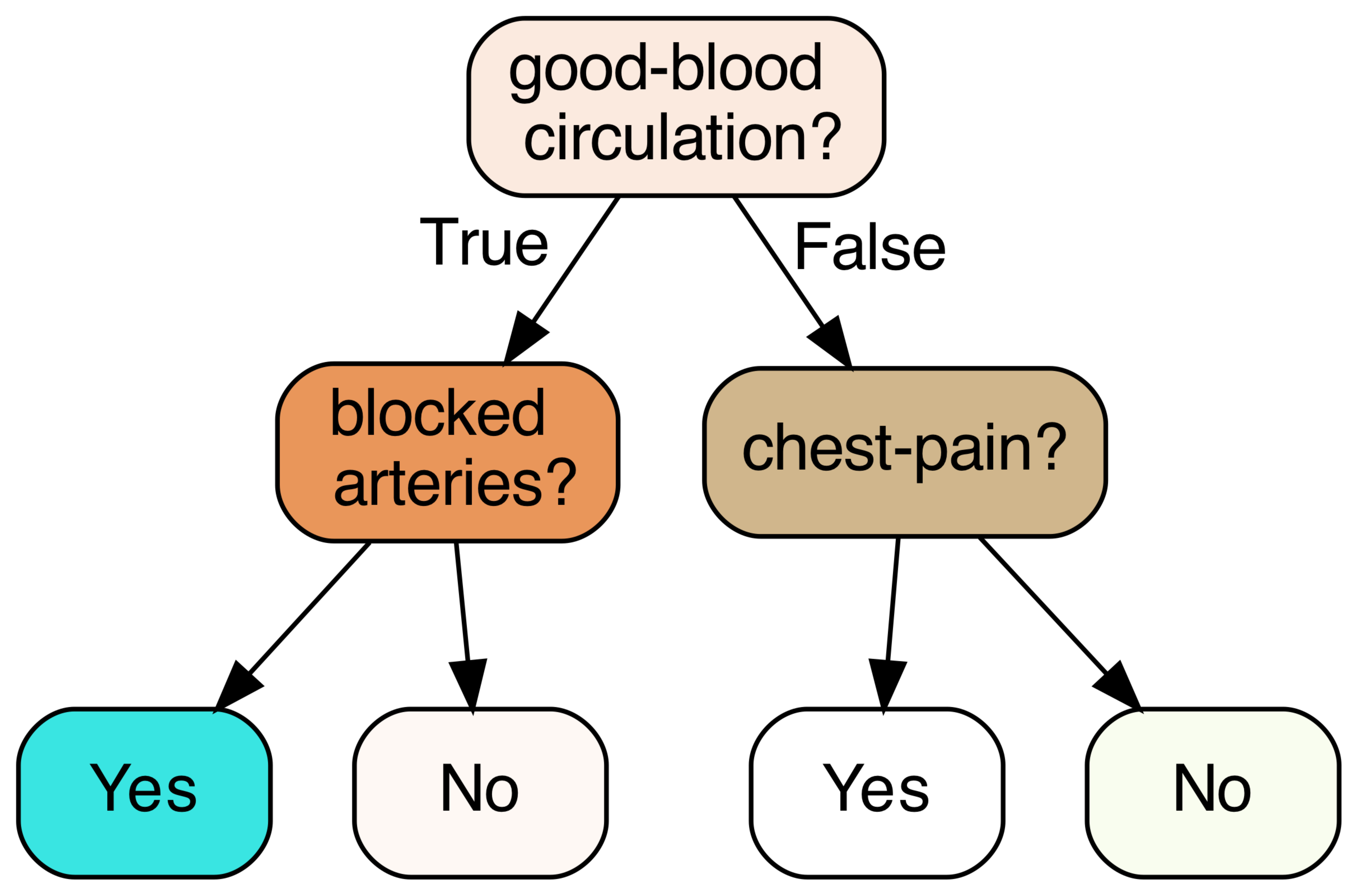}}
\end{minipage}   
\caption{Running example.}
 \label{fig:RE}
\end{figure}

\paragraph{Running Example.}
 Let us assume a simple binary classification problem for  predicting 
 whether  or not a patient has a heart disease. 
 The class variables are: {\it Yes} and {\it No} 
 ({\it Yes} to classify the patient as suffering from heart disease and {\it No} 
 to classify the patient  as  without heart disease.)  
 and a set of features in the following order:  {\it blocked-arteries},  
 {\it good-blood-circulation},  {\it chest-pain}, and {\it weight},  
 where features 1, 2 and 3  
 represent Boolean variables, and  feature 4 represents an ordinal variable. 
 Let the set of trees, shown in Figure \ref{fig:RE}, be the tree ensemble of 
 an RF classifier $\mathfrak{F}$ trained on the heart disease problem and 
 $\tau$ its classification function. 
 There are 3 trees in the forest and each tree has a maximum depth of 2.  
 Assume we have an instance $\mbf{v}=(1, 0, 1, 70)$, namely, 
 {\it blocked-arteries} = 1, {\it good-blood-circulation} = 0, 
  {\it chest-pain} = 1,  {\it weight} = 70. 
 Hence, Trees 1 and 3 vote for {\it Yes} and  Tree 2 votes for {\it No}.  
 As the majority votes go for {\it Yes}, then the classifier will return 
  {\it Yes} for  $\mbf{v}$, i.e. $\tau ( \mbf{v} )  =  \textit{Yes}$.  
  
 \paragraph{Boolean satisfiability (SAT).}
 The paper assumes the notation and definitions standard in
 SAT~\cite{sat21}, i.e.\ the decision problem for propositional logic,
 which is known to be NP-complete~\cite{Cook71}.
A propositional formula $\varphi$ is defined over a finite set of
Boolean variables 
$X = \{x_1, x_2, \ldots, x_n\}$. 
Formulas are most often represented in \emph{conjunctive normal form} (CNF). 
A CNF formula is a conjunction of clauses, a clause is a
disjunction of literals, and a literal is a variable ($x_i$) or its
negation ($\neg{x_i}$).
Whenever convenient, a formula is viewed as a set of sets of literals. 
A Boolean interpretation $\mu$ of a formula $\varphi$ is a total mapping
of $X$ to $\{0,1\}$ 
($0$ corresponds to \emph{False} and $1$ corresponds to \emph{True}).
Interpretations can be extended to literals, clauses and formulas with
the usual semantics; hence we can refer to $\mu(l)$, $\mu(\omega)$,
$\mu(\varphi)$, to denote respectively the value of a literal, clause
and formula given an interpretation.
Given a formula $\varphi$,  $\mu$ is a \emph{model} of $\varphi$ 
if it makes $\varphi$ \emph{True}, i.e. $\mu(\phi) = 1$. 
A formula $\varphi$ is \emph{satisfiable} ($\varphi \nentails \perp$)
if it admits a model,
otherwise,  it is \emph{unsatisfiable}  ($\varphi \entails \perp$). 
Given two formulas $\varphi$ and $\psi$, we say that $\varphi$ {\it
  entails} $\psi$ (denotes $\varphi\entails\psi$) if all models of
$\varphi$ are also models of $\psi$. 
$\varphi$ and $\psi$ are equivalent (denoted $\varphi \equiv \psi$) if
$\varphi \entails \psi$ and $\psi \entails \varphi$. 

\paragraph*{Abductive explanations.}
The paper uses the definition of {\it PI-explanation} \cite{ShihCD18} 
(also referred to as {\it abductive explanation} (AXp) in \cite{IgnatievNM19})
\footnote{Throughout the paper we will use both terms {\it PI-explanation} 
and {\it abductive explanation} (AXp) interchangeably.}, 
based on prime implicants of
some decision function (related with the predicted class).
Let us consider a given ML model, computing a classification function
$\tau$ on feature space $\mbb{F}$, a point $\mbf{v}\in\mbb{F}$, with
prediction $c=\tau(\mbf{v})$, with $\mbf{v}=(v_1,\ldots,v_m)$. 
%
A PI-explanation (AXp) is any minimal subset $\fml{X}\subseteq\fml{F}$
such that:
\begin{equation} \label{eq:pixpl}
  \forall(\mbf{x}\in\mbb{F}).
  \left[
    \bigwedge\nolimits_{i\in{\fml{X}}}(x_i=v_i)
    \right]
  \limply(\tau(\mbf{x})=c)
\end{equation}

In a similar vein, we consider contrastive explanations (CXps) 
\cite{miller-aij19}. A formal definition of contrastive explanation is 
proposed in recent work \cite{inams-aiia20}.

\paragraph*{Contrastive explanations.}
Contrastive explanation can be defined as a minimal subset  
$\fml{Y}\subseteq\fml{F}$ that suffice to changing the prediction 
if features of $\fml{Y}$ are allowed to take some arbitrary value from their domain. 
Given $\mbf{v}=(v_1,\ldots,v_m)\in\mbb{F}$ with $\kappa(\mbf{v})=c$, a
CXp is any minimal subset $\fml{Y}\subseteq\fml{F}$ such that,
\begin{equation} \label{eq:cxp}
  \exists(\mbf{x}\in\mbb{F}).\bigwedge\nolimits_{j\in\fml{F}\setminus\fml{Y}}(x_j=v_j)\land(\kappa(\mbf{x})\not=c) 
\end{equation}
Building on the results of R.~Reiter in model-based
diagnosis~\cite{reiter-aij87},~\cite{inams-aiia20} proves a minimal
hitting set (MHS) duality relation between AXps and CXps,
i.e.\ AXps are MHSes of CXps and vice-versa.

\begin{example}
Consider the binary classifier $\mathfrak{F}$ of the running example. 
and the instance $(\mbf{v}=(1,0,1,70),\mathit{Yes})$.
If the features {\it good-blood-circulation} and {\it weight} are
allowed to take any possible value from their domain, and the values of
the features {\it blocked-arteries} and {\it chest-pain} are kept to
their values in $\mbf{v}$, then the prediction is still {\it Yes}.
This means that the features {\it good-blood-circulation} and 
{\it weight} can be deemed irrelevant for the prediction of {\it Yes}
given the other feature values in $\mbf{v}$.
Moreover, by allowing either {\it blocked-arteries} or {\it
  chest-pain} to take any value, prediction will change to
{\it No}. 
Hence, $\set{\textit{blocked-arteries},\textit{chest-pain}}$ is a 
subset-minimal set of features sufficient for predicting  
$\tau(\mbf{v})=\textit{Yes}$, that is a PI-explanation (AXp). 
Additionally, setting the value of  $\textit{blocked-arteries}$ to 0 suffices 
to changing the prediction of  $\mbf{v}$  (i.e. $\tau(0,0,1,70)=\textit{No}$),
thus $\set{\textit{blocked-arteries}}$ is a CXp.
\end{example}

\section{Complexity of AXps for RFs}
\label{sec:cplx}

Recent work identified classes of classifiers for which one
AXp can be computed in polynomial
time~\cite{marquis-kr20,msgcin-nips20}. These classes of classifiers
include those respecting specific criteria of the knowledge
compilation map~\cite{marquis-kr20}\footnote{%
  The knowledge compilation map was first proposed in
  2002~\cite{darwiche-jair02}.},
but also Naive Bayes and linear classifiers (resp.~NBCs and
LCs)~\cite{msgcin-nips20}. 
(In the case of NBCs and LCs, enumeration of AXps was shown
to be solved with polynomial delay.)
One question is thus whether there might exist a polynomial time
algorithm for computing one computing AXp of an RF.
This section shows that this is unlikely to be the case, by proving
that deciding whether a set of features represents an AXp is
$\ddp$-complete\footnote{The class $\ddp$ \cite{Papadimitriou94} is
  the set of languages defined by the intersection of two languages,
  one in NP and one in coNP.}. 

Let $\mathfrak{F}$ be an RF, with classification function $\tau$, and
let $\mbf{v}\in\mbb{F}$, with prediction $\tau(\mbf{v})=c\in\fml{K}$.
$\tau$ is parameterized with $c$, to obtain the boolean function
$\tau_c$, s.t.\ $\tau_c(\mbf{x})=1$ iff $\tau(\mbf{x})=c$.
A set of literals $I_{\mbf{v}}$ is associated with each $\mbf{v}$.
Let $\rho$ be a subset of the literals associated with $\mbf{v}$,
i.e.\ $\rho\subseteq{I_{\mbf{v}}}$. Hence,

\begin{theorem}
  For a random forest $\mathfrak{F}$, given an instance $\mbf{v}$ with
  prediction $c$, deciding whether a set of literals is an
  AXp is $\ddp$-complete.
\end{theorem}

\begin{Proof}
  Given an instance $\mbf{v}$ and predicted class $c$, deciding
  whether a set of literals $\rho$ is an AXp of an RF
  $\mathfrak{F}$ is clearly in $\ddp$. We need to prove that
  $\rho\entails\tau_c$, which is a problem in coNP. We also need to
  prove that a set of literals $\rho'$, obtained by the removal of any
  single literal from $\rho$ (and there can be at most $m$ of these),
  is such that $\rho\nentails\tau_c$, a problem in NP.
  To prove that the problem is hard for $\ddp$,
  we reduce the problem of computing a prime implicant of a DNF, which
  is known to be complete for $\ddp$~\cite{umans-tcad06}, to the
  problem of computing a PI-explanation of an RF $\mathfrak{F}$.\\
  Consider a DNF $\phi=t_1\lor\cdots\lor{t_n}$, where each term $t_i$
  is a conjunction of literals defined on a set $X=\{x_1,\ldots,x_m\}$
  of boolean variables. 
  Given $\phi$, we construct an RF $\mathfrak{F}$, defined on a set
  $\fml{F}$ of $m$ features, where each feature $i$ is associated with
  an $x_i$ element of $X$, and where $D_i=\{0,1\}$. Moreover,
  $\mathfrak{F}$ is such that  $\phi(\mbf{x})=1$ iff
  $\tau_1(\mbf{x})=1$.
  $\mathfrak{F}$ is constructed as follows.
  \begin{enumerate}[nosep,label=\roman*.]
  \item
    Associate a decision tree (DT) $\fml{T}_i$ with each term
    $t_i$, such that the assignment satisfying $t_i$ yields class 1,
    and the other assignments yield class 0. Clearly, the size of the
    DT $\fml{T}_i$ is linear on the size of $t_i$, since each literal
    not taking the value specified by the term will be connected to a
    terminal node with prediction 0.
    \item
    Create $(n-1)$ additional trees, each having exactly one
    terminal node and no non-terminal nodes. Moreover, associate class
    $1$ with the terminal node.
    \end{enumerate}
  Next, we prove that $\phi(\mbf{x})=1$ iff $\tau_1(\mbf{x})=1$.
  \begin{enumerate}[nosep,label=\roman*.]
  \item
    Let $\mbf{x}$ be such that $\phi(\mbf{x})=1$. Then, there is
    at least one term $t_j$, such that $t_j(\mbf{x})=1$. As a
    result, the corresponding tree $\fml{T}_j$ in the RF will
    predict class 1.  Hence, at least $n$ trees predict class 1, and
    at most $n-1$ trees predict class 0. As a result, the predicted
    class is 1, and so $\tau_1(\mbf{x})=1$.
    \item
    Let $\mbf{x}$ be such that $\tau_1(\mbf{x})=1$. This means
    that at least one of the trees associated with the terms $t_i$
    must predict value 1. Let such tree be $\fml{T}_j$, associated
    with term $t_j$. For this tree to predict class 1, then
    $t_j(\mbf{x})=1$, and so $\phi(\mbf{x})=1$.
    \end{enumerate}
  Now, let $\rho$ be a conjunction of literals defined on $X$. Then,
  we must have $\rho\entails\phi$ iff $\rho\entails\tau_1$. Every model
  of $\rho$ is also a model of $\phi$, and so it must also be a model
  of $\tau_1$. Conversely, every model of $\rho$ is also a model of
  $\tau_1$, and so it must also be a model of $\phi$.
\end{Proof}

\section{AXps for Random Forests}
\label{sec:enc}

This section outlines the computation of PI-explanations (AXps) for RFs.
We first present the algorithm's organization. The algorithm requires a
logical encoding of RFs, which are presented next.

\paragraph{Computing AXps.}

A minimal set of features $\fml{X}\in\fml{F}$ is an AXp
if~\eqref{eq:pixpl} holds. Clearly, this condition holds iff the
following formula is unsatisfiable,
\[
\left[\bigwedge\nolimits_{i\in{\fml{X}}}(x_i=v_i)\right]
\land
\msf{Enc}(\tau(\mbf{x})\not=c)
\]
The previous formula has two components
$\langle\fml{H},\fml{S}\rangle$. 
$\fml{H}$ represents the set of hard clauses, encoding the
representation of the ML model and also imposing a constraint on
the predicted class, i.e.\ $\msf{Enc}(\tau(\mbf{x})\not=c)$.
%
$\fml{S}$ represents the unit (\emph{soft}) clauses, each capturing a
literal $(x_i=v_i)$. Since the clauses in $\fml{S}$ are \emph{soft},
they can be \emph{dropped} (thus allowing $x_i$ to take
any value) while searching for a minimal subset of $\fml{E}$ of
$\fml{S}$, such that,
\[
\left[\bigwedge\nolimits_{(x_1=v_i)\in\fml{E}}(x_i=v_i)\right]
\land
\msf{Enc}(\tau(\mbf{x})\not=c)
\]
is unsatisfiable. Our goal is to find a minimal set $\fml{S}$ such
that the pair $\langle\fml{H},\fml{S}\rangle$ remains
unsatisfiable (where $\fml{S}$ can be viewed as the background
knowledge against which the clauses in $\fml{S}$ are inconsistent).
This corresponds to finding a minimal unsatisfiable subset (MUS) of
$\langle\fml{H},\fml{S}\rangle$,
%
and so any off-the-shelve MUS extraction algorithm can be used
for computing an AXp (as noted in earlier
work~\cite{IgnatievNM19}).

Clearly, adapting the descibed procedure above of computing one AXp to 
one that computes a CXp is straightforward.  That is, the minimal set $\fml{Y}$
of $\fml{S}$ to search is, such that,
\[
\left[\bigwedge\nolimits_{(x_1=v_i)\in\fml{S}\setminus\fml{Y}}(x_i=v_i)\right]
\land
\msf{Enc}(\tau(\mbf{x})\not=c)
\]
is satisfiable. Further, hitting set duality between  AXps and CXps allows  
to exploit any algorithm for computing MUSes/MCSes\footnote{An MCS is a minimal 
  set of clauses to remove from an unsatisfiable CNF formula to recover
  consistency . It is well-known that MCSes are minimal hitting sets
  of MUSes and vice-versa~\cite{reiter-aij87,lozinskii-jetai03}.} to
enumerate both kinds of explanations (AXps and CXps).  (Recent
work~\cite{inams-aiia20,DL:icml21,DL:sat21a} exploits the MUS/MCS
enumeration algorithm MARCO~\cite{lpmms-cj16} for enumerating
AXps/CXps.)

We detail next how to encode an RF, while requiring some prediction
not to hold. We start with a simple encoding of an RF into an SMT 
formula, and then we detail a purely propositional encoding, which
enables the use of SAT solvers.

\paragraph{SMT Encoding.}
~
Several encodings of tree ensemble models, such as  Boosted Trees (BTs), 
have been proposed and they are essentially based on SMT/MILP 
(see e.g. \cite{ChenZS0BH19,EinzigerGSS19,Ignatiev20}, etc). 
Hence, it is natural to follow prior work 
and propose a straightforward encoding of RFs in SMT. 
Intuitively, the formulation of RFs into SMT formulas is as follows.
We encode every single DT of an RF as a set of  implication rules.
That is, a DT path (classification rule) is interpreted as a rule of 
the form  $\textit{antecedent} \limply \textit{consequent}$ where 
the \textit{antecedent} is a conjunction of predicates 
encoding the non-terminal nodes of the path and the       
\textit{consequent} is a  predicate representing 
the class associated with the terminal node of the path. 
Next, we aggregate the prediction (votes) of the DTs and count the prediction 
score for each class. This can be expressed by an arithmetic function that calculates 
the sum of trees predicting the same class. Lastly, a linear inequality 
checks which class has the largest score. 

The implementation of the encoding above resulted in performance
results comparable to those of BTs \cite{Ignatiev20}. However, in the
case of RFs it is possible to devise a purely propositional encoding,
as detailed below.

\paragraph{SAT Encoding.}
~
Our goal is to represent the structure of an RF with a propositional
formula. This requires abstracting away what will be shown to be
information used by the RF that is unnecessary for finding an
AXp.
Concretely, and as shown below, the actual values of the features used
as literals in the RF need not be considered when computing one
AXp.
We start by detailing how to encode the nodes of the decision trees in
an RF. This is done such that the actual values of the features are
abstracted away. Then, we present the encoding of the RF classifier
itself, including how the majority voting is represented.

To encode a terminal node of a DT, we proceed as follows.
Given a set of classes  $\fml{K} = \{c_1, c_2, \ldots, c_K \}$, a
terminal node $t$ labeled with one class of $\fml{K}$. Then, 
we define for each $c_j \in{\fml{K}}$ a variable $l_j$ and represent 
the terminal node $t$ with its corresponding label class $c_j$, i.e. $\kappa(t) = c_j$.

Moreover, the encoding of a non-terminal node of a DT is organized as
follows.
Given a feature $j \in{\fml{F}}$ associated with a non-terminal node $l$, 
with a domain $\mbb{D}_j$, each outgoing edge of $l$ is represented by 
a literal $l_j$ of the form  $x_j \bowtie S_j$ s.t. $x_j \in{\fml{F}}$ is the variable 
of  feature $j$, $S_j \bowtie \mbb{D}_j$ and $\bowtie\:\in{\set{=, \subseteq, \in}}$.  
Hence we distinguish three cases for encoding $x_j \bowtie S_j$.
For the first case, feature $j$ is binary, and so the literal
$l_j$ is True if  $x_j = 1$ and False if $x_j = 0$.
For the second case, feature $j$ is categorical, and so we introduce a
Boolean variable $z_i$  
for each value $v_i \in \mbb{D}_j$ s.t.\ $z_i = 1$ iff $x_j = v_i$. 
Assume $S_j =\set{v_1, \ldots, v_n}$, then we connect $l_j$ to variables $z_i, i = 1, \ldots, n$  
as follows: $l_j \lequiv (z_1 \vee \ldots \vee z_n)$ or  $\neg l_j \lequiv (z_1 \vee \ldots \vee z_n)$, 
depending on whether the current edge is going to left or right child-node. 
Finally, for the third case, feature $j$ is real-valued. 
Thus, $S_j$ is either an interval or a union of intervals.
Concretely, we  consider 
all the splitting thresholds of the feature $j$ existing in the  RF and we generate 
(in order) all the possible intervals. Each interval $I_i  \subseteq \mbb{D}_j$ is denoted 
by a Boolean variable $z_i$.  Let us assume $S_j = I_1 \cup \ldots \cup I_n$ and 
$z_1, \ldots, z_n$ are variables associated with   $I_1 \cup \ldots \cup I_n$, then 
we have $l_j \lequiv (z_1 \vee \ldots \vee z_n)$ or $\neg l_j \lequiv (z_1 \vee \ldots \vee z_n)$. 
Moreover, this encoding can be reduced into a more compact set of
constraints  (that also improves propagation in the SAT solver).
Indeed, if the number of intervals is large the encoding will be as
well. Hence, we propose to use auxiliary variables in the encoding.
Assume $\mbb{D}_j = I_1 \cup I_2 \cup I_3 $, nodes $l$ and $l^{\prime}$
s.t. $\neg l_j \lequiv z_1$, $\neg l_j \lequiv (z_1 \vee z_2 \vee z_3)$, 
$\neg l_j^{\prime} \lequiv (z_1 \vee z_2)$, $ l_j^{\prime} \lequiv z_3$, then 
this can be re-written as:  
 $\neg l_j \lequiv z_1$,  $l_j \lequiv (z_2 \vee l_j^{\prime})$, 
 $\neg l_j^{\prime} \lequiv (z_2 \vee \neg l_j)$, $l_j^{\prime} \lequiv z_3$.
The next step is to encode the RF classifier.
Given an RF $\mathfrak{F}$ formed by a set of $M$ DTs, i.e.  
$\mathfrak{F} = \set{\fml{T}_1, \fml{T}_2, \ldots, \fml{T}_M }$, 
and $\tau$ the classification function of $\mathfrak{F}$.  
For every DT $ \fml{T}_i \in  \mathfrak{F}$,  we encode the  set of paths 
 $\fml{R}_i$ of $\fml{T}_i$ as follows:
\begin{equation}
\bigwedge\nolimits_{R_k \in{\fml{R}_i}}
\left(\left( 
 \bigwedge\nolimits_{l_j \in\fml{L}(R_k)} l_j
\right)
\limply l_{ij}
\right)
 \label{eq:path}    
\end{equation}
where $l_{ij}$ is a literal associated with $\fml{T}_i$ in which 
the voted class is $c_j \in{\fml{K}}$. 
For every  DT $\fml{T}_i \in  \mathfrak{F}$, and every path $R_k \in{\fml{R}_i}$, 
we enforce the condition that exactly one variable  $l_{ij}$ is {\it True} in 
$\fml{T}_i$ \footnote{Only one class is returned by  $\fml{T}_i $.}. 
This can be expressed as\footnote{%
  We use the well-known cardinality networks~\cite{asin2011} for
  encoding all the cardinality constraints of the proposed encoding.}:
\begin{equation} 
\left(\bigvee\nolimits_{j=1}^{K} l_{ij}\right)  ~ \wedge ~  \sum\nolimits_{j=1}^{K} l_{ij} \leq 1
\label{eq:oneClass}
\end{equation}
%
%


Finally, for capturing the majority voting used in RFs, we need to
express the constraint that the counts (of individual tree selections)
for some class $c_k\not=c_j$ have to be large enough (i.e.\ greater
than, or greater than or equal to) when compared to the counts for
class $c_j$.
We start by showing how cardinality constraints can be used for
expressing such constraints. The proposed solution requires the use of
$K-1$ cardinality constraints, each comparing the counts of $c_j$ with
the counts of some other $c_k$.
Afterwards, we show how to reduce to 2 the number of cardinality
constraints used.

Let $c_j\in\fml{K}$ be the predicted class. The index $j$ is relevant
and ranges from 1 to $K=|\fml{K}|$.
Moreover, let $1\le{k}<{j}\le{K}$.
Class $c_k$ is selected over class $c_j$ if:
\begin{equation} \label{eq:be01}
  \small
  \sum\nolimits_{i=1}^{M}l_{ik}>\sum\nolimits_{i=1}^{M}l_{ij}
  \Leftrightarrow
  \sum\nolimits_{i=1}^{M}l_{ik}+\sum\nolimits_{i=1}^{M}\neg{l_{ij}}\ge{M}
\end{equation}
Similarly, let $1\le{j}<{k}\le{K}$.
Class $c_k$ is selected over class $c_j$ if:
\begin{equation} \label{eq:be02}
  \small
  \sum\nolimits_{i=1}^{M}l_{ik}>\sum\nolimits_{i=1}^{M}l_{ij}
  \Leftrightarrow
  \sum\nolimits_{i=1}^{M}l_{ik}+\sum\nolimits_{i=1}^{M}\neg{l_{ij}}\ge{M+1} 
\end{equation}
(A simple observation is that these constraints can be optimized for
the case $|\fml{K}|=2$.)

It is possible to reduce the number of cardinality constraints as
follows.
Let us represent \eqref{eq:be01} and \eqref{eq:be02}, respectively, by
$\msf{Cmp}_{\prec}(z^{\prec}_1,\ldots,z^{\prec}_M)$ and
$\msf{Cmp}_{\succ}(z^{\succ}_1,\ldots,z^{\succ}_M)$. (Observe that the
encodings of these constraints differ (due to the different RHSes).)
Moreover, in~\eqref{eq:be01} and~\eqref{eq:be02} we replace 
 $(z^{\prec}_1,\ldots,z^{\prec}_M)$ and
$(z^{\succ}_1,\ldots,z^{\succ}_M)$ resp.\ by $l_{1k},\ldots,l_{Mk}$,
for some $k$. The idea is that we will \emph{only} use two
cardinality constraints, one for \eqref{eq:be01} and one for
\eqref{eq:be02}.

Let $p_k=1$ iff $k$ is to be compared with $j$.
In this case,
$\msf{Cmp}_{\bowtie}(z^{\bowtie}_1,\ldots,z^{\bowtie}_M)$ (where
$\bowtie$ is either $\prec$ or $\succ$) is comparing the class counts
of $c_j$ with the class counts of some $c_k$.
Let us use the constraint $p_k\limply(z^{\bowtie}_i\leftrightarrow{l_{ik}})$,
with $k\in\{1,\ldots,{K}\}\setminus\{j\}$, and
$1\le{i}\le{M}$. This serves to allow associating the (free) variables
$(z^{\bowtie}_1,\ldots,z^{\bowtie}_M)$ with some set of literals
$(l_{1k},\ldots,l_{Mk})$. 
Moreover, we also let $p_0\limply(z^{\bowtie}_i\leftrightarrow{1})$ and 
$p_j\limply(z^{\bowtie}_i\leftrightarrow{1})$, i.e.\ we allow $p_0$
and $p_j$ to pick \emph{guaranteed winners}, and such that
$\neg{p_0}\lor\neg{p_j}$.
Essentially, we allow for a guaranteed winner to be picked below $j$
or above $j$, but not both.
Clearly, we must pick one $k$, either below or above $j$, to pick a
class $c_k$, when comparing the counts of $c_k$ and $c_j$.
We do this by picking two winners, one below $j$ and one above $j$,
and such that at most one is allowed to be a guaranteed winner (either
0 or $j$):
\[
\left(\sum\nolimits_{r=0}^{j-1}p_r=1\right)\:\land\:\left(\sum\nolimits_{r=j}^{K}p_r=1\right)
\]
Observe that, by starting the sum at 0 and $j$, respectively, we allow
the picking of one guaranteed winner in each summation, if needed be.
To illustrate the  described  encoding above we consider again the running example.
\begin{example}
Consider again the RF $\mathfrak{F}$ of the running example and the instance 
$\mbf{v} = (1, 0, 1, 70)$, giving the prediction {\it Yes}. 
Let us define the Boolean  variables $x_1, x_2$ and $x_3$ associated with 
the binary features  {\it blocked-arteries},   {\it good-blood-circulation}, 
{\it chest-pain}, resp. and variables $w_1$ and $w_2$  representing   
$(\textit{weight} > 75)$ and $(\textit{weight} \le 75)$ resp. and an auxiliary variable $w$. 
 Also, to represent the classes {\it No} and {\it Yes},  
 we associate  
 variables denoting the classes for each tree:  
 $\set{l_{11}, l_{12}}$ for $\fml{T}_1$,    
 $\set{l_{21}, l_{22}}$ for $\fml{T}_2$  and 
 $\set{l_{31}, l_{32}}$ for $\fml{T}_3$.    
 Hence, the corresponding set of encoding constraints is:
 $\{ x_1 \wedge x_3 \limply l_{12}$,
 $ x_1 \wedge \neg x_3 \limply l_{11}$,
 $\neg x_1 \limply l_{11}$, 
 $x_2 \limply l_{21}$, 
 $\neg x_2 \wedge w \limply l_{22}$, 
 $\neg x_2 \wedge \neg w \limply l_{21}$,
 $x_2 \wedge x_1 \limply l_{32}$,
 $x_2 \wedge \neg x_1 \limply l_{31}$,
 $\neg x_2 \wedge x_3 \limply l_{32}$,
 $\neg x_2 \wedge \neg x_3 \limply l_{31}$,
 $w \lequiv w_1$,
 $\neg w \lequiv w_2$,
 $(l_{11} + l_{21} + l_{31}) \ge 2$,
 $x_1$, $\neg x_2 $,  $x_3$,  $w_2 \}$.
 Observe that $\{ x_1, \neg x_2 ,  x_3,  w_2 \}$ denotes the set of
 the soft  constraints whereas the remaining are the hard
 constraints.  
\end{example}

\section{Experimental Results}
\label{sec:res}

\setlength{\tabcolsep}{5pt}
\rowcolors{2}{gray!10}{}
\newcommand{\lpr}{(}
\newcommand{\rpr}{)}

\begin{table*}[h]
\centering
\resizebox{\textwidth}{!}{
  \begin{tabular}{l>{\lpr}S[table-format=3.0,table-space-text-pre=\lpr]cS[table-format=2.0,table-space-text-post=\rpr]<{\rpr}cS[table-format=4.0]cS[table-format=5.0]S[table-format=5.0]ccS[table-format=1.0]S[table-format=1.0]cccS[table-format=3.0] | cS[table-format=1.0]}
\toprule[1.2pt]
\rowcolor{white}
\multirow{2}{*}{\bf Dataset} & \multicolumn{3}{c}{\multirow{2}{*}{\bf (\#F~~~~\#C~~~~\#I)}}  & \multicolumn{3}{c}{\bf RF} & \multicolumn{2}{c}{\bf CNF}  & \multicolumn{4}{c}{\bf SAT oracle} & \multicolumn{4}{c}{\bf AXp (\texttt{RFxpl})}  &  \multicolumn{2}{c}{\bf Anchor}\\
  \cmidrule[0.8pt](lr{.75em}){5-7}
  \cmidrule[0.8pt](lr{.75em}){8-9}
  \cmidrule[0.8pt](lr{.75em}){10-13}
  \cmidrule[0.8pt](lr{.75em}){14-17}
  \cmidrule[0.8pt](lr{.75em}){18-19} 
\rowcolor{white}
& \multicolumn{3}{c}{} & {\bf D}  & {\bf \#N} & {\bf \%A} & {\bf \#var} & {\bf \#cl} &  {\bf MxS} & {\bf MxU} & {\bf \#S} & {\bf \#U}  & {\bf Mx } & {\bf m }  & {\bf avg } & {\bf \%w} & {\bf avg} & {\bf \%w} \\
\toprule[1.2pt]

ann-thyroid & 21 & 3 & 718 & 4 & 2192 & 98 & 17854 & 29230 & 0.12 & 0.15 & 2 & 18 & 0.36 & 0.05 & 0.13 & 96 & 0.32 &4 \\
appendicitis & 7 & 2 & 43 & 6 & 1920 & 90 & 5181 & 10085 & 0.02 & 0.02 & 4 & 3 & 0.05 & 0.01 & 0.03 & 100 & 0.48 & 0\\
banknote & 4 & 2 & 138 & 5 & 2772 & 97 & 8068 & 16776 & 0.01 & 0.01 & 2 & 2 & 0.03 & 0.02 & 0.02 & 100 & 0.19 & 0 \\
biodegradation & 41 & 2 & 106 & 5 & 4420 & 88 & 11007 & 23842 & 0.31 & 1.05 & 17 & 22 & 2.27 & 0.04 & 0.29 & 97 & 4.07 & 3 \\
ecoli & 7 & 5 & 66 & 5 & 3860 & 90 & 20081 & 34335 & 0.10 & 0.09 & 4 & 2 & 0.21 & 0.05 & 0.10 & 98 & 0.38 & 2 \\
glass2 & 9 & 2 & 66 & 6 & 2966 & 90 & 7303 & 15194 & 0.04 & 0.03 & 5 & 4 & 0.09 & 0.02 & 0.03 & 100 & 0.61 & 0 \\
heart-c & 13 & 2 & 61 & 5 & 3910 & 85 & 5594 & 11963 & 0.04 & 0.02 & 6 & 7 & 0.07 & 0.01 & 0.04 & 100 & 0.85 & 0 \\
ionosphere & 34 & 2 & 71 & 5 & 2096 & 87 & 7174 & 14406 & 0.02 & 0.02 & 22 & 11 & 0.11 & 0.02 & 0.03 & 100 & 12.43 &0 \\
iris & 4 & 3 & 60 & 6 & 1446 & 93 & 16346 & 25603 & 0.02 & 0.01 & 2 & 2 & 0.05 & 0.02 & 0.03 & 100 & 0.21 & 0 \\
karhunen & 64 & 10 & 200 & 5 & 6198 & 91 & 36708 & 70224 & 1.06 & 1.41 & 35 & 29 & 14.64 & 0.65 & 2.78 & 100 & 28.15& 0 \\
letter & 16 & 26 & 398 & 8 & 44304 & 82 & 28991 & 68148 & 1.97 & 3.31 & 8 & 8 & 6.91 & 0.24 & 1.61 & 70 & 2.48 & 30 \\
magic & 10 & 2 & 381 & 6 & 9840 & 84 & 29530 & 66776 & 0.51 & 1.84 & 6 & 4 & 2.13 & 0.07 & 0.14 & 99 & 0.91 & 1 \\
mofn-3-7-10 & 10 & 2 & 128 & 6 & 8776 & 92 & 2926 & 8646 & 0.00 & 0.01 & 3 & 7 & 0.02 & 0.01 & 0.01 & 100 & 0.29 & 0\\
new-thyroid & 5 & 3 & 43 & 5 & 1766 & 100 & 17443 & 28134 & 0.03 & 0.01 & 3 & 2 & 0.08 & 0.03 & 0.05 & 100 & 0.36 & 0 \\
pendigits & 16 & 10 & 220 & 6 & 12004 & 95 & 30522 & 59922 & 2.40 & 1.32 & 10 & 6 & 4.11 & 0.14 & 0.94 & 96 & 3.68 &4 \\
phoneme & 5 & 2 & 540 & 6 & 8962 & 82 & 21899 & 49840 & 0.09 & 0.07 & 3 & 2 & 0.22 & 0.05 & 0.09 & 98 & 0.37 & 2 \\
ring & 20 & 2 & 740 & 6 & 6188 & 89 & 19114 & 42362 & 0.27 & 0.44 & 11 & 9 & 1.25 & 0.05 & 0.25 & 92 & 7.25 & 8 \\
segmentation & 19 & 7 & 42 & 4 & 1966 & 90 & 21288 & 35381 & 0.11 & 0.17 & 8 & 10 & 0.53 & 0.11 & 0.31 & 100 & 4.13 & 0 \\
shuttle & 9 & 7 & 1160 & 3 & 1460 & 99 & 18669 & 29478 & 0.11 & 0.08 & 2 & 7 & 0.34 & 0.05 & 0.14 & 99 & 0.42 & 1 \\
sonar & 60 & 2 & 42 & 5 & 2614 & 88 & 9938 & 20537 & 0.04 & 0.06 & 36 & 24 & 0.43 & 0.04 & 0.09 & 100 & 23.02 & 0 \\
spambase & 57 & 2 & 442 & 5 & 4614 & 92 & 13055 & 28284 & 0.07 & 0.06 & 18 & 37 & 0.50 & 0.05 & 0.11 & 100 & 6.18 & 0 \\
spectf & 44 & 2 & 54 & 5 & 2306 & 88 & 6707 & 13449 & 0.07 & 0.06 & 20 & 24 & 0.34 & 0.02 & 0.07 & 100 & 8.12 & 0 \\

texture & 40 & 11 & 550 & 5 & 5724 & 87 & 34293 & 64187 & 0.79 & 0.63 & 23 & 17 & 3.24 & 0.19 & 0.93 & 100 & 28.13 &0 \\
 
threeOf9 & 9 & 2 & 103 & 3 & 920 & 100 & 2922 & 4710 & 0.00 & 0.00 & 1 & 8 & 0.01 & 0.00 & 0.01 & 100 & 0.14 & 0 \\
 
twonorm & 20 & 2 & 740 & 5 & 6266 & 94 & 21198 & 46901 & 0.08 & 0.08 & 12 & 8 & 0.28 & 0.06 & 0.10 & 100 & 5.73 & 0 \\
 
vowel & 13 & 11 & 198 & 6 & 10176 & 90 & 44523 & 88696 & 1.66 & 2.11 & 8 & 5 & 4.52 & 0.15 & 1.15 & 66 & 1.67 & 34 \\
 
waveform-21 & 21 & 3 & 500 & 5 & 6238 & 84 & 29991 & 57515 & 0.68 & 0.54 & 10 & 11 & 2.27 & 0.09 & 0.34 & 100 & 5.86& 0 \\
 
waveform-40 & 40 & 3 & 500 & 5 & 6232 & 83 & 30438 & 58380 & 0.50 & 0.86 & 15 & 25 & 7.07 & 0.11 & 0.88 & 100 & 11.93 & 0 \\
 
wdbc & 30 & 2 & 114 & 4 & 2028 & 96 & 7813 & 15742 & 0.06 & 0.02 & 12 & 18 & 0.13 & 0.03 & 0.05 & 100 & 10.56 & 0 \\
 
wine-recog & 13 & 3 & 72 & 3 & 1188 & 97 & 17718 & 28421 & 0.04 & 0.04 & 5 & 8 & 0.13 & 0.04 & 0.07 & 100 & 1.46 & 0 \\
 
wpbc & 33 & 2 & 78 & 5 & 2432 & 76 & 9078 & 18675 & 1.00 & 1.53 & 20 & 13 & 5.33 & 0.03 & 0.65 & 79 & 3.91 & 21 \\
 
xd6 & 9 & 2 & 172 & 6 & 8288 & 100 & 2922 & 8394 & 0.00 & 0.00 & 3 & 6 & 0.02 & 0.01 & 0.01 & 100 & 0.57 & 0 \\

\bottomrule[1.2pt]
\end{tabular}
}
\caption{%
  Performance evaluation of the RF explainability tool (\texttt{RFxpl}),
  and comparison with Anchor. 
The table shows results for 32 datasets, i.e.\ those for which test
data accuracy is no less than 76\%.
Columns {\bf\#F}, {\bf\#C} and {\bf\#I} report, respectively, the
number of features, classes and tested instances, in the dataset.
Columns {\bf D}, {\bf\#N} and {\bf\%A} show, respectively, each tree's
max.~depth, total number of nodes and test accuracy of an RF classifier.
Columns  {\bf \#var} and {\bf \#cl} show the number of variables and 
clauses of a CNF formula encoding an RF classifier along with any
instance to analyze.
Column {\bf MxS}  (resp.~{\bf MxU}) reports the maximum runtime of a 
SAT call (UNSAT call, resp.) and column  {\bf \#S} ({\bf \#U}, resp.) reports 
the average number of SAT calls (resp.~UNSAT calls) performed to extract  
an AXp. 
Column {\bf avg} ({\bf Mx} and {\bf m}, resp.) shows the average 
(max. and min., resp.) runtime for extracting an explanation.
%
The percentage of won instances is given as {\bf \%w}.    
}
\label{tab:res}
\end{table*}

This section assess the performance of our approach to compute 
PI-explanations (AXps) for RFs  and also compares the results with an heuristic 
explaining model Anchor \cite{Ribeiro0G18}\footnote{%
  Anchor computes heuristic \emph{local} explanations and not
  PI-explanations (AXp). Also, we do not consider other tools, such as 
  LIME \cite{Ribeiro0G16} or SHAP \cite{LundbergL17}, as these
  learn a simpler ML model as an explanation and not a set of literals.}
The assessment is performed on a selection of 32 publicly available 
datasets, which originate from {\it UCI Machine Learning Repository}~\cite{uci} 
and {\it Penn Machine Learning Benchmarks}~\cite{Olson2017PMLB}.
Benchmarks comprise  binary and multidimensional classification datasets. 
The number of classes in the benchmark suite varies from 2 to 26.   
The number of features (resp.~data samples)  varies from 4 to 64 
(106 to 58000, resp.) with the average being 21.9 (resp.~5131.09). 
When training RF classifiers for the selected datasets, we used 80\% of
the dataset instances (20\% used for test data). For assessing 
explanation tools, we randomly picked fractions of the dataset,
depending on the dataset size. Concretely, for datasets containing, resp.,  
less than 200, 200\textendash999, 1000\textendash 9999 and 10000 or 
more, we randomly picked, resp., 40\%, 20\%, 10\% and 2\% of the
instances to analyze.  
The number of trees in each RF is set to 100 while tree depth  
varies between 3 and 8. (Note that we have tested other values for 
the number of trees ranging from 30 to 200, and we fixed it to 100
since with 100 trees RFs achieve the best train/test accuracies.) 
As a result, the accuracy of the trained models varies between 76\% 
to 100\%.
We use the scikit-learn \cite{scikit-learn} 
ML tool to train RF models.  Note that, scikit-learn can only handle binary and 
ordinal features in the case of RFs.  Accordingly,  
the experiments focus on binary and continuous data and 
do not include categorical features.  
In addition, we have overridden the implemented RF learner 
in sciki-learn so that it reflects  the original algorithm described in \cite{Breiman01}
\footnote{%
The RF model implemented by scikit-learn uses probability estimates  
to predict a class, whereas in the original proposal for
RFs~\cite{Breiman01}, the prediction is based on majority vote.}.
Furthermore, PySAT \cite{imms-sat18} is used to instrument incremental SAT oracle calls.

The experiments are conducted on a MacBook Pro with a Dual-Core Intel
Core~i5 2.3GHz CPU with 8GByte RAM running macOS Catalina.
\autoref{tab:res} summarizes the results of assessing the performance 
of our RF explainer tool (\texttt{RFxpl}) on the selected datasets.
(The table's caption also describes the meaning of each column.)
As can be observed, and with three exceptions, the average running
time of \texttt{RFxpl} is less than 1sec.\ per instance. In terms of
the largest running times, there are a few outliers (this is to be
expected since we are solving a $\ddp$-hard problem), and these occur
when the number of classes is large.
To assess the scalability of \texttt{RFxpl}, we compared
\texttt{RFxpl} with
the well-known heuristic explainer Anchor~\cite{Ribeiro0G18}.
(Clearly, the exercise does not aim to compare the explanation
accuracies of Anchor and \texttt{RFxpl}, but only to assess how
scalable it is in practice to solve a $\ddp$-complete explainability
problem with a propositional encoding and a SAT oracle.)
Over all datasets, \texttt{RFxpl} outperforms Anchor on more than 96\%
of the instances (i.e.\ 8438 out of 8746). In terms of average running
time per instance, \texttt{RFxpl} outperforms Anchor by more than 1
order of magnitude, concretely the average run time of Anchor is 14.22
times larger than the average runtime of \texttt{RFxpl}.


\section{Conclusion}
\label{sec:conc}

This paper proposes a novel approach for explaining random forests.
First, the paper shows that it is $\ddp$-complete to decide whether a
set of literals is a PI-explanation (AXp) for a random forest.
Second, the paper proposes a purely propositional encoding for
computing PI-explanations (AXps) of random forests.
The experimental results allow demonstrating that the proposed
approach not only scales to realistically sized random forests, but it
is also significantly more efficient that a state of the art
model-agnostic heuristic approach.
Given the practical interest in RFs~\cite{ZhouF17}, finding
AXps represents a promising new application area of SAT
solvers.

Two extensions of the work can be envisioned. One extension is
to improve further the propositional encoding proposed in this
paper, aiming at eliminating the very few cases where heuristic
approaches are more efficient.
Another extension is to exploit the proposed model (and any of its
future improvements) to explain deep random forests~\cite{ZhouF17}.


\paragraph{Acknowledgements.}
This work is supported by the AI Interdisciplinary Institute ANITI,
funded by the French program ``Investing for the Future - PIA3'' under
Grant agreement n$^{o}$ ANR-19-PI3A-0004.

\bibliographystyle{namedabbrv}
\bibliography{refs}

\begin{thebibliography}{}

\bibitem[\protect\citeauthoryear{As{\'\i}n \bgroup \em et al.\egroup
  }{2011}]{asin2011}
R.~As{\'\i}n, R.~Nieuwenhuis, A.~Oliveras, and E.~Rodr{\'\i}guez-Carbonell.
\newblock Cardinality networks: a theoretical and empirical study.
\newblock {\em Constraints}, 16(2):195--221, 2011.

\bibitem[\protect\citeauthoryear{Audemard \bgroup \em et al.\egroup
  }{2020}]{marquis-kr20}
G.~Audemard, F.~Koriche, and P.~Marquis.
\newblock On tractable {XAI} queries based on compiled representations.
\newblock In {\em {KR}}, pages 838--849, 2020.

\bibitem[\protect\citeauthoryear{Biere \bgroup \em et al.\egroup
  }{2021}]{sat21}
A.~Biere, M.~Heule, H.~van Maaren, and T.~Walsh, editors.
\newblock {\em Handbook of Satisfiability}. {IOS} Press, 2021.

\bibitem[\protect\citeauthoryear{Birnbaum and
  Lozinskii}{2003}]{lozinskii-jetai03}
E.~Birnbaum and E.~L. Lozinskii.
\newblock Consistent subsets of inconsistent systems: structure and behaviour.
\newblock {\em J. Exp. Theor. Artif. Intell.}, 15(1):25--46, 2003.

\bibitem[\protect\citeauthoryear{Breiman \bgroup \em et al.\egroup
  }{1984}]{BreimanFOS84}
L.~Breiman, J.~H. Friedman, R.~A. Olshen, and C.~J. Stone.
\newblock {\em Classification and Regression Trees}.
\newblock Wadsworth, 1984.

\bibitem[\protect\citeauthoryear{Breiman}{2001}]{Breiman01}
L.~Breiman.
\newblock Random forests.
\newblock {\em Mach. Learn.}, 45(1):5--32, 2001.

\bibitem[\protect\citeauthoryear{Chen \bgroup \em et al.\egroup
  }{2019}]{ChenZS0BH19}
H.~Chen, H.~Zhang, S.~Si, Y.~Li, D.~S. Boning, and C.~Hsieh.
\newblock Robustness verification of tree-based models.
\newblock In {\em NeurIPS}, pages 12317--12328, 2019.

\bibitem[\protect\citeauthoryear{Cook}{1971}]{Cook71}
S.~A. Cook.
\newblock The complexity of theorem-proving procedures.
\newblock In {\em STOC}, pages 151--158, 1971.

\bibitem[\protect\citeauthoryear{Darwiche and Hirth}{2020}]{darwiche-ecai20}
A.~Darwiche and A.~Hirth.
\newblock On the reasons behind decisions.
\newblock In {\em ECAI}, pages 712--720, 2020.

\bibitem[\protect\citeauthoryear{Darwiche and Marquis}{2002}]{darwiche-jair02}
A.~Darwiche and P.~Marquis.
\newblock A knowledge compilation map.
\newblock {\em J. Artif. Intell. Res.}, 17:229--264, 2002.

\bibitem[\protect\citeauthoryear{Einziger \bgroup \em et al.\egroup
  }{2019}]{EinzigerGSS19}
G.~Einziger, M.~Goldstein, Y.~Sa'ar, and I.~Segall.
\newblock Verifying robustness of gradient boosted models.
\newblock In {\em {AAAI}}, pages 2446--2453, 2019.

\bibitem[\protect\citeauthoryear{Feng and Zhou}{2018}]{FengZ18a}
J.~Feng and Z.~Zhou.
\newblock {AutoEncoder} by forest.
\newblock In {\em {AAAI}}, pages 2967--2973, 2018.

\bibitem[\protect\citeauthoryear{Gao and Zhou}{2020}]{GaoZ20}
W.~Gao and Z.~Zhou.
\newblock Towards convergence rate analysis of random forests for
  classification.
\newblock In {\em NeurIPS}, 2020.

\bibitem[\protect\citeauthoryear{Guidotti \bgroup \em et al.\egroup
  }{2019}]{GuidottiMRTGP19}
R.~Guidotti, A.~Monreale, S.~Ruggieri, F.~Turini, F.~Giannotti, and
  D.~Pedreschi.
\newblock A survey of methods for explaining black box models.
\newblock {\em {ACM} Comput. Surv.}, 51(5):93:1--93:42, 2019.

\bibitem[\protect\citeauthoryear{Ignatiev and
  Marques{-}Silva}{2021}]{DL:sat21a}
A.~Ignatiev and J.~Marques{-}Silva.
\newblock {SAT}-based rigorous explanations for decision lists.
\newblock In {\em {SAT}}, 2021.
\newblock In Press.

\bibitem[\protect\citeauthoryear{Ignatiev \bgroup \em et al.\egroup
  }{2018}]{imms-sat18}
A.~Ignatiev, A.~Morgado, and J.~Marques{-}Silva.
\newblock {PySAT:} {A} {Python} toolkit for prototyping with {SAT} oracles.
\newblock In {\em SAT}, pages 428--437, 2018.

\bibitem[\protect\citeauthoryear{Ignatiev \bgroup \em et al.\egroup
  }{2019}]{IgnatievNM19}
A.~Ignatiev, N.~Narodytska, and J.~Marques{-}Silva.
\newblock Abduction-based explanations for machine learning models.
\newblock In {\em {AAAI}}, pages 1511--1519, 2019.

\bibitem[\protect\citeauthoryear{Ignatiev \bgroup \em et al.\egroup
  }{2020}]{inams-aiia20}
A.~Ignatiev, N.~Narodytska, N.~Asher, and J.~Marques{-}Silva.
\newblock From contrastive to abductive explanations and back again.
\newblock In {\em AI*IA}, 2020.
\newblock (Preliminary version available from
  \url{https://arxiv.org/abs/2012.11067}.).

\bibitem[\protect\citeauthoryear{Ignatiev}{2020}]{Ignatiev20}
A.~Ignatiev.
\newblock Towards trustable explainable {AI}.
\newblock In {\em {IJCAI}}, pages 5154--5158, 2020.

\bibitem[\protect\citeauthoryear{Li \bgroup \em et al.\egroup }{2018}]{LiLCR18}
O.~Li, H.~Liu, C.~Chen, and C.~Rudin.
\newblock Deep learning for case-based reasoning through prototypes: {A} neural
  network that explains its predictions.
\newblock In {\em AAAI}, pages 3530--3537, 2018.

\bibitem[\protect\citeauthoryear{Liffiton \bgroup \em et al.\egroup
  }{2016}]{lpmms-cj16}
M.~H. Liffiton, A.~Previti, A.~Malik, and J.~Marques{-}Silva.
\newblock Fast, flexible {MUS} enumeration.
\newblock {\em Constraints An Int. J.}, 21(2):223--250, 2016.

\bibitem[\protect\citeauthoryear{Lundberg and Lee}{2017}]{LundbergL17}
S.~M. Lundberg and S.~Lee.
\newblock A unified approach to interpreting model predictions.
\newblock In {\em NeurIPS}, pages 4765--4774, 2017.

\bibitem[\protect\citeauthoryear{Marques{-}Silva \bgroup \em et al.\egroup
  }{2020}]{msgcin-nips20}
J.~Marques{-}Silva, T.~Gerspacher, M.~C. Cooper, A.~Ignatiev, and
  N.~Narodytska.
\newblock Explaining naive bayes and other linear classifiers with polynomial
  time and delay.
\newblock In {\em NeurIPS}, 2020.

\bibitem[\protect\citeauthoryear{Marques{-}Silva \bgroup \em et al.\egroup
  }{2021}]{DL:icml21}
J.~Marques{-}Silva, T.~Gerspacher, M.~C. Cooper, A.~Ignatiev, and
  N.~Narodytska.
\newblock Explanations for monotonic classifiers.
\newblock In {\em {ICML}}, 2021.
\newblock In Press.

\bibitem[\protect\citeauthoryear{Miller}{2019}]{miller-aij19}
T.~Miller.
\newblock Explanation in artificial intelligence: Insights from the social
  sciences.
\newblock {\em Artif. Intell.}, 267:1--38, 2019.

\bibitem[\protect\citeauthoryear{Mollas \bgroup \em et al.\egroup
  }{2020}]{MollasBVT20}
I.~Mollas, N.~Bassiliades, I.~P. Vlahavas, and G.~Tsoumakas.
\newblock Lionforests: local interpretation of random forests.
\newblock In {\em ECAI}, volume 2659, pages 17--24, 2020.

\bibitem[\protect\citeauthoryear{Montavon \bgroup \em et al.\egroup
  }{2018}]{MontavonSM18}
G.~Montavon, W.~Samek, and K.~M{\"{u}}ller.
\newblock Methods for interpreting and understanding deep neural networks.
\newblock {\em Digit. Signal Process.}, 73:1--15, 2018.

\bibitem[\protect\citeauthoryear{Olson \bgroup \em et al.\egroup
  }{2017}]{Olson2017PMLB}
R.~S. Olson, W.~La~Cava, P.~Orzechowski, R.~J. Urbanowicz, and J.~H. Moore.
\newblock Pmlb: a large benchmark suite for machine learning evaluation and
  comparison.
\newblock {\em BioData Mining}, 10(1):36, 2017.

\bibitem[\protect\citeauthoryear{Papadimitriou}{1994}]{Papadimitriou94}
C.~H. Papadimitriou.
\newblock {\em Computational complexity}.
\newblock Addison-Wesley, 1994.

\bibitem[\protect\citeauthoryear{Pedregosa \bgroup \em et al.\egroup
  }{2011}]{scikit-learn}
F.~Pedregosa, G.~Varoquaux, A.~Gramfort, V.~Michel, B.~Thirion, O.~Grisel,
  M.~Blondel, P.~Prettenhofer, R.~Weiss, V.~Dubourg, J.~Vanderplas, A.~Passos,
  D.~Cournapeau, M.~Brucher, M.~Perrot, and E.~Duchesnay.
\newblock Scikit-learn: Machine learning in {P}ython.
\newblock {\em Journal of Machine Learning Research}, 12:2825--2830, 2011.

\bibitem[\protect\citeauthoryear{Petkovic \bgroup \em et al.\egroup
  }{2018}]{PetkovicAWV18}
D.~Petkovic, R.~B. Altman, M.~Wong, and A.~Vigil.
\newblock Improving the explainability of random forest classifier - user
  centered approach.
\newblock In {\em PSB}, pages 204--215, 2018.

\bibitem[\protect\citeauthoryear{Quinlan}{1993}]{Quinlan93}
J.~R. Quinlan.
\newblock {\em {C4.5:} Programs for Machine Learning}.
\newblock Morgan Kaufmann, 1993.

\bibitem[\protect\citeauthoryear{Reiter}{1987}]{reiter-aij87}
R.~Reiter.
\newblock A theory of diagnosis from first principles.
\newblock {\em Artif. Intell.}, 32(1):57--95, 1987.

\bibitem[\protect\citeauthoryear{Ribeiro \bgroup \em et al.\egroup
  }{2016}]{Ribeiro0G16}
M.~T. Ribeiro, S.~Singh, and C.~Guestrin.
\newblock "why should {I} trust you?": Explaining the predictions of any
  classifier.
\newblock In {\em KDD}, pages 1135--1144, 2016.

\bibitem[\protect\citeauthoryear{Ribeiro \bgroup \em et al.\egroup
  }{2018}]{Ribeiro0G18}
M.~T. Ribeiro, S.~Singh, and C.~Guestrin.
\newblock Anchors: High-precision model-agnostic explanations.
\newblock In {\em {AAAI}}, pages 1527--1535, 2018.

\bibitem[\protect\citeauthoryear{Shih \bgroup \em et al.\egroup
  }{2018}]{ShihCD18}
A.~Shih, A.~Choi, and A.~Darwiche.
\newblock A symbolic approach to explaining bayesian network classifiers.
\newblock In {\em {IJCAI}}, pages 5103--5111, 2018.

\bibitem[\protect\citeauthoryear{Shih \bgroup \em et al.\egroup
  }{2019}]{ShihCD19}
A.~Shih, A.~Choi, and A.~Darwiche.
\newblock Compiling bayesian network classifiers into decision graphs.
\newblock In {\em {AAAI}}, pages 7966--7974, 2019.

\bibitem[\protect\citeauthoryear{UCI}{2020}]{uci}
{UCI Machine Learning Repository}.
\newblock \url{https://archive.ics.uci.edu/ml}, 2020.

\bibitem[\protect\citeauthoryear{Umans \bgroup \em et al.\egroup
  }{2006}]{umans-tcad06}
C.~Umans, T.~Villa, and A.~L. Sangiovanni{-}Vincentelli.
\newblock Complexity of two-level logic minimization.
\newblock {\em {IEEE} Trans. Comput. Aided Des. Integr. Circuits Syst.},
  25(7):1230--1246, 2006.

\bibitem[\protect\citeauthoryear{Yang \bgroup \em et al.\egroup
  }{2020}]{YangWJZ20}
L.~Yang, X.~Wu, Y.~Jiang, and Z.~Zhou.
\newblock Multi-label learning with deep forest.
\newblock In {\em {ECAI}}, volume 325, pages 1634--1641, 2020.

\bibitem[\protect\citeauthoryear{Zhang \bgroup \em et al.\egroup
  }{2019}]{Zhang19}
Y.~Zhang, J.~Zhou, W.~Zheng, J.~Feng, L.~Li, Z.~Liu, M.~Li, Z.~Zhang, C.~Chen,
  X.~Li, Y.~A. Qi, and Z.~Zhou.
\newblock Distributed deep forest and its application to automatic detection of
  cash-out fraud.
\newblock {\em {ACM} Trans. Intell. Syst. Technol.}, 10(5):55:1--55:19, 2019.

\bibitem[\protect\citeauthoryear{Zhao \bgroup \em et al.\egroup
  }{2019}]{ZhaoWLC19}
X.~Zhao, Y.~Wu, D.~L. Lee, and W.~Cui.
\newblock iforest: Interpreting random forests via visual analytics.
\newblock {\em {IEEE} Trans. Vis. Comput. Graph.}, 25(1):407--416, 2019.

\bibitem[\protect\citeauthoryear{Zhou and Feng}{2017}]{ZhouF17}
Z.~Zhou and J.~Feng.
\newblock Deep forest: Towards an alternative to deep neural networks.
\newblock In {\em {IJCAI}}, pages 3553--3559, 2017.

\end{thebibliography}

\end{document}